# k-NS: Section-Based Outlier Detection in High Dimensional Space

Zhana

## ABSTRACT

Finding rare information hidden in a huge amount of data from the Internet is a necessary but complex issue. Many researchers have studied this issue and have found effective methods to detect anomaly data in low dimensional space. However, as the dimension increases, most of these existing methods perform poorly in detecting outliers because of "high dimensional curse". Even though some approaches aim to solve this problem in high dimensional space, they can only detect some anomaly data appearing in low dimensional space and cannot detect all of anomaly data which appear differently in high dimensional space. To cope with this problem, we propose a new k-nearest section-based method (k-NS) in a section-based space. Our proposed approach not only detects outliers in low dimensional space with section-density ratio but also detects outliers in high dimensional space with the ratio of k-nearest section against average value. After taking a series of experiments with the dimension from 10 to 10000, the experiment results show that our proposed method achieves 100% precision and 100% recall result in the case of extremely high dimensional space, and better improvement in low dimensional space compared to our previously proposed method.

## Categories and Subject Descriptors

H.2.8 [**Database Management**]: Database Applications- Data mining. I.5.3 [Pattern Recognition]: Outlier Detection.

## 1. INTRODUCTION

Seeking for meaningful information from very large database is always a significant issue in data mining field. These valued data are called anomaly data, which are different from the rest of normal data based on some measures. They are also called outliers from a distance or density view. Many definitions about outliers are proposed according to different perspectives. The widely accepted definition about outlier is Hawkins': an outlier is an observation that deviates so much from other observations as to arouse suspicion that it was generated by a different mechanism [7]. This definition not only describes the difference of data from observation, but also points out the essential difference of data in mechanism. In the low dimensional space, outliers can be considered as far points from the normal points based on the distance. However, in high dimensional space, the distance doesn't meet the exact description between the anomaly data and normal data any longer. The only distinguished point is the difference in distributions between normal data and anomaly data. Most of existing methods insist on finding the outliers by distance even in high dimensional space. They can find anomaly data obviously far from normal data, but ignore the anomaly data that are inside the range of normal data. Therefore, we give a new definition about anomaly data in high dimensional space in our approach: "most of the data conform to one distribution such as the normal distribution, while the small part of data conforms to another different distribution or just distribute randomly. These rare data are called anomaly data and only can be detected by some specific measures".

In this paper, we reconsider the concept of outliers, explain the section space concept, and then propose a new algorithm called k-NS. The main features and contributions of this paper are summarized as follows:

- We analyze the connection of data distribution between the different dimensions. By this, the high dimensional problem is transformed into the dimensional-loop problem. This problem is easily solved by some statistic method.

- Our proposed method uses the k-NN (k Nearest Neighbor) concept for the section calculation and puts forward to a novel k-NS (k Nearest Sections) concept. Hence, our proposal can better evaluate the disperse degree of points with neighbor points projected in a new dimension.

- We reconsider the concept of outlier, and employ section space instead of traditional space in order to avoid the "high dimensional curse" problem.

- We execute a series of experiments with the range of dimensions from 10 to 10000 to evaluate our proposed algorithm. The experiment results clearly show that our proposed algorithm has significant advantages over other algorithms with stably and precisely in large volumes of data when the dimension increases extremely high. Even in the 10,000 dimensional data experiment, our algorithm easily achieves 100% precision and 100% recall result compared with ever-proposed anomaly detection algorithms.

- We also point out the difference between the outliers and noisy data. The outliers are obviously different in high dimensional space with noisy data. These two concepts are confusing because they are mixed together in low dimensional space and even considered as identical by some researchers.

- Furthermore, we analyze the feature of data distribution in high dimensional space. The study on high dimensional space provides a new view to solve the anomaly data detection issue.

This paper is organized as follows. In section 2, we give a brief overview of related works on high dimensional outlier detection. In section 3, we introduce our concept and our novel approach, and we also describe our proposal and discuss some optimizations. In section 4, we evaluate the proposed method by experiments of different dimensional artificial dataset and real dataset. And we conclude the findings in section 5.

## 2. RELATED WORKS

As an important sub-tree of the data mining field, anomaly data detection has been developed for more than ten years, and many study results have been achieved in large scale database. We categorize them into the following five groups to introduce these methods clearly.

**Distance and Density Based Outlier Detection:** The distance based outlier detection is a classic method because it comes from the original outlier definition, i.e. Outliers are those points that are far from other points based on distance measures, e.g. by Hilout[8]. This algorithm detects point with its k-nearest neighbors by distance and uses space-filling curve to map high dimensional space. The most well-known LOF [1] uses k-NN and density based algorithm, which detects the outlier locally by its k-nearest distance neighbor points and measures outliers by lrd (local reachability density) and LOF (Local Outlier Factor). This algorithm runs smoothly in low dimensional space and is still effective in relative high dimensional space. LOCI [9] is an improved algorithm based on LOF, which is more sensitive to local distance than LOF. However, LOCI performs worse than LOF in high dimensional space.

**Subspace Clustering Based Outlier Detection:** Since it is difficult to find outliers in high dimensional space, they try to find these points behaving abnormally in low dimensional space. Subspace clustering is a feasible method applied to outlier detection in high dimensional space. This approach assumes that outliers are always deviated from others in low dimensional space if they are different in high dimensional space. Aggarwal [2] uses the equi-depth ranges in each dimension with expected fraction and deviation of points in k-dimensional cube D given by $N \times f^k$ and $\sqrt{N \times f^k \times (1 - f^k)}$. This method detects outliers by calculating the sparse coefficient S(D) of the cube D.

**Outlier Detection with Dimension Deduction:** Another method is dimension deduction from high dimensional space to low dimensional space, such as SOM (Self-Organizing Map) [18] [19], mapping several dimensions to two dimensions, and then detecting the outliers in two dimensional space. FindOut [11] detects outliers by removing the clusters and deducts dimensions with wavelet transform on multidimensional data. However, this method may cause information loss when the dimension is reduced. The result is not as robust as expected, and then it is seldom applied to outlier detection.

**Information-theory based Outlier Detection:** The distribution of points in each dimension can be coded for data compression, hence high dimensional issue changes to information statistic issue in each dimension. Christian Bohm has proposed CoCo [3] method with MDL(Minimum Description Length) for outlier detection, and he also applies this method to the clustering issue, e.g. Robust Information-theory Clustering[5][12].

**Other Outlier Detection Methods:** Besides above four groups, some detection measurements are also distinctive and useful. One notable approach is called ABOD (Angle-Based Outlier Detection) [4]. It is based on the concept of angle with vector product and scalar product. The outliers usually have the smaller angles than normal points.

These methods have reduced the "high dimensional cures" moderately and get the correct results in some special cases. However, "high dimensional curse" problem still exists and affects the point's detection. Christian Bohm's information-based method is similar to the subspace clustering methods and is only applied to detect outliers in low dimensional space.

In summary, seeking a general approach to detect outliers in high dimensional space is still a key issue that needs to be solved.

## 3. PROPOSED METHOD

It is well known that Euclidean distance between points in high dimensional space becomes obscure and immeasurable. That is why there is "high dimensional curse". Moreover, outlier detection in subspace or dimension reduction may cause the information lost or only valid on specific dataset. We still need to find a suitable way for outlier detection in high dimensional space.

### 3.1 General Idea

Learning from the subspace outlier detection methods, we know that high dimensional issue can be transformed into the statistical issue by loop detection in different subspaces. We have also noticed that points' positions change differently in different dimensions. By observation and analysis of these points, we have found that the outliers are placed in a big cluster of normal points in some dimensions and deviated from these points in other dimensions. Otherwise in another situation, the outliers are clustered differently in different dimensions from normal points while normal points are always clustered together in all dimensions. Therefore, our proposed method only needs to solve the issue with two conditions: whether there are points with low density in low dimensional space; or whether there are points that are deviated from other points in the same section of one dimension when these points are projected to other dimensions.

Our proposal can be divided into four steps. First, we divide the entire range of data into small regions in each dimension. Here, we call the small region a section. Based on the section divisions, we construct the new data structure called section space, which is different from traditional Euclidean space. Second, we compare point's scattering with others in different sections in each dimension by computing the section density value. Third, we compare the point's dispersing in the same section of one dimension after projecting it to other dimensions. Last, we sum up all results for each point, and then compare these points with a statistic measure. The outliers are the points whose values are obviously higher than most of the points.

### 3.2 Section Data Structure

Our proposed method is based on the section data structure. The mechanism on how to compose this section structure and transform the Euclidean data space into our proposed section space is necessarily introduced in detail in below.

We divide the space into the same number of sections in each dimension, so the space just looks like a grid. The conventional data space information is composed of points and dimensions while our proposed data structure represents the data distribution with point, dimension and section. This structure overcomes the shortcomings of distance measurement in conventional space, and it is easy to observe the distribution changes of points in one section while sections and dimensions are varied in different situations.

The data structure is described by the PointInfo (point information) and SectionInfo (section information) as follows:

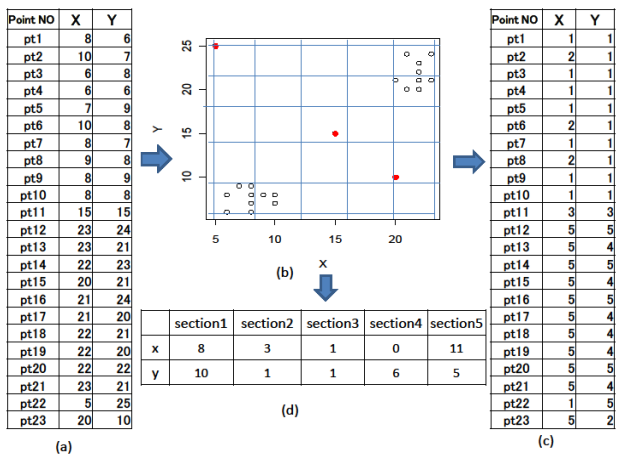

**Figure 1:** Section Space Division and Dimension Projection

---

*PointInfo*[Dimension ID, Point ID]: section ID of the *point*

*SectionInfo*[Dimension ID, Section ID]: #points in the section

---

An example of two dimensional data to explain the transforming process from original dataset to a new section space is shown in Figure 1.

The example dataset includes 23 points in two dimensions as shown in Figure 1(a). The original data are placed into the data space as in the conventional method, as shown in Figure 1(b). In our proposed section-based structure, we construct the PointInfo structure as illustrated in Figure 1(c) and SectionInfo as shown in Figure 1(d). The range of each dimension is divided into five sections. The section division sample is shown in data space with blue lines as in Figure 1(b) and SectionInfo as in Figure 1(d).

We notice that the range of each dimension is different. If we set the largest range for all dimensions, there must be too many blank sections in the small range dimensions and the blank sections produce meaningless values 0 would affect the result markedly in following calculations. Therefore, we set the range from minimum coordinate to the maximum coordinate for each dimension. In order to avoid two end sections having larger density than that of other sections, we loose the border by enlarge the original range by 0.1%. Taking the data in Figure 1 as an example to explain how to generate the range of the data in each dimension. The original range in *x* dimension is (5, 23), and the length is 18. The new range is (4.991, 23.009) by enlarging the length by 0.1%, therefore, the new length is18.018. The original range in *y* dimension is (6, 25), and the length is 19. The new range is (5.9905, 25.0095), and the new length is 19.019. The length of section is 3.6036 in *x* dimension and 3.8038 in *y* dimension.

## 3.3 Definitions

To provide an explicit explanation for our proposed method, some definitions are formulated as follows:

| Symbol | Definition |
|---|---|
| **P** (point) | The information of point. $p_j$ refers to the $j^{th}$ point of all points. $p_{i,j}$ refers to the $j^{th}$ point in $i^{th}$ dimension. |
| **Section** (section) | The range of data in each dimension is divided into the same number of equi-width parts, which are called sections. The width of the section is determined by the section density and the range in that dimension. |
| *scn* (number of section) | The number of sections for each dimension. It is decided by the number of total points and the average section density. *scn* is defined equally in each dimension. |
| *d* (section density) | The number of points in one section is called section density. $d_i$ means the average section density in $i^{th}$ dimension. $d_{i,j}$ means the $j^{th}$ section density of $i^{th}$ dimension. $d_i(p)$ means the section density of the point *p* to in $i^{th}$ dimension. |
| *dists* (section distance) | The section distance used for evaluating the section difference among points in all projected dimensions, as defined in Equation (4). |
| *SecVal* (point section value) | A point deviation value in each dimension as defined in Equation (3). |
| *SecValp* (projected point section value) | A point deviation value in projected dimensions as defined in Equation (5). |
| *SI* (statistic information) | The statistic information of each point as defined in Equation (6). |

*d* is the section density, which presents different meanings in different cases. Case1: in one section, all points of this section have the same section density, and $d_{i,j}$ means section density value in the $i^{th}$ dimension and $j^{th}$ section. All points in this section of dimension i have the same $d_{i,j}$ value. Case2: the section density is used to compare with the average value in this dimension. So the low section density has a low ratio with the average section density in one dimension. $d_i$ means the average section density in $i^{th}$ dimension. Case3: if the section density of a point is needed, the expression should include the point. $d_i(p)$ means the section density value of point *p* in $i^{th}$ dimension. From above on, different subscriptions express a specific meaning for point, section of dimension, although they are all expressed by *d*.

Dimension projection is a concept that needs to be clarified. Each point has its coordinate value in different dimensions. All dimensions have the equal relations. However, when we observe the change of points in different dimensions, we need to fix the points of the same section in a certain dimension at first, and then compare the distance change of these points in other different dimensions. The initial dimension is called original dimension, and the different dimensions are called projected dimension. It means that we project the points from the original dimension to the other dimensions.

## 3.4 k-NS

The outlier detection method in section space is different from the method in a conventional data space. The existing methods such as those using distance and density among points cannot be applied in the section space directly. In section space, each point gets the values in different sections and dimensions, and then all these values decide whether the points are outliers or not. We propose a novel method effective in high dimensional space, called k-nearest sections or k-NS. This proposed method is a statistical approach which detects outliers in each dimension and the projected dimensions.

Before we introduce k-NS definition, the *dists* (section distance) need to be clarified in advance. It is noted that the *section distance* defined in this paper is different from the general definition of distance.

**DEFINITION 1** (*dists* of points)

Let point $p, q \in section$. $p_i, q_i$ are in $i^{th}$ dimension. When $p_i, q_i$ are projected from dimension i to j, the section distance between them corresponds to the difference of their section ID.

$$dists(p_i, q_i) = |SecId(p_j) - SecId(q_j)| + 1 \quad (1)$$

In original dimension *i*, p and *q* are in the same section. After the dimension change from i to j, *p* and *q* have a new section ID. So we can compare the section difference between $SecId(p_j)$ and $SecId(q_j)$. The *dists* is the absolute difference value between any two points. The minimum *dists* of point is 1, which is different from the conventional Euclidean distance concept. Because the frequency of points in the same section is more than that of points with the same coordinate, there must be many minimum sections distance. If all neighbor points have minimum *dists* 0, the point gets value 0 from neighbor point's *dists*. In this case, it easily cause illegal computation and invalid to compare with other points. So the minimum *dists* is set 1 to solve this question perfectly.

The k-NS mathematic definition is based on the *dists*, which can find outliers in different situations.

**DEFINITION 2** (k-NS)

The $x_{ns}$ of a given point $x \in Section$ in the database $D \subset \mathbb{R}^d$ is defined as

$$x_{ns} = \{x, x' \in D | \forall x \in D, x, x', p \in Section, \sum_{i=1}^{m} d_i(x) << \sum_{i=1}^{m} d_i \cup \sum_{i=1}^{m}\sum_{j=1, j\neq i}^{m} dists(x', p) \leq \sum_{i=1}^{m}\sum_{j=1, j\neq i}^{m} dists(x, p)\} \quad (2)$$

*x*, *x`* and *p* are the points in the same section in dimension *i*. The point *p* is anyone of k-nearest points which give *dists* of neighbor points to compare it with other points from the original same section when projecting them to new dimensions. k-NS gives the effective method with *dists* to discriminate the points of the same section by projecting the dimension from original to others d. $x_{ns}$ is a statistical value for summarizing all values of point in each dimension and in all projected dimension, which means the $x_{ns}$ could be used as a whole result finding outliers and impossible to be found in some certain dimensions.

From the k-NS definition, outliers should satisfy either of two conditions: The first condition guarantees that outliers can be detected in low dimensional space; the second condition guarantees that outliers still can be detected with neighbor points' *dists* in the projected dimensions if the point does not appear abnormal in low dimensional space.

Our proposed method uses loop calculation in dimensions instead of the calculation of the Euclidean distance. Therefore, each point needs to be evaluated in each dimension and each projected dimensions on first step.

1. Section Density Ratio Comparison in Each Dimension

Outlier points always appear more sparsely than most normal points if they can be detected in low dimensions. Therefore, the density of outliers is lower than the average density in that dimension. The section density ratio of points with the average section density does not only reflect the sparsity in the dimension, but also compare points between different dimensions.

**DEFINITION 3** (Section Density Ratio)

Set point $p_{i,j} \in Section_{i,k}$ in dimension *i*, where *j* is point ID and *k* is section ID. $d_{i,k}$ is section density of point $p_{i,j}$ in dimension *i* and $d_i$ is the average section density in dimension *i*. The point's *SecVal* in each dimension is defined as follows:

$$SecVal(p_{i,j}) = \left(\frac{d_{i,k}}{d_i}\right)^2 \quad (3)$$

It is to be noticed that *SecVal* is not the section density value, but a ratio of the section density with the average value in that dimension. The reason is that a point has different sparsity in different dimensions. If points only use section density to compare, the value is easily affected in different dimensions and cannot get accurate result. The ratio of section density overcomes this shortcoming, which is independent among dimensions and only decided by the distribution of the points in its own dimension. Another noticed aspect is that one *SecVal* does not only correspond to one point, but it actually presents the section's density value. Hence the points which are in the same section share the same *SecVal* values.

2: k-Nearest Section Comparison in Each Projected Dimensions

If the outliers don't appear farther in the low dimensions, they cannot be detected by the first step. Since they hide among the normal points and have similar distance or density with others. Nevertheless, these points still can be found different from other points in the same section in the projected dimensions. This step aims to find outliers from normal points by projecting these points into different dimensions. The section distance measurement is a effective method to compare these points. Based on the section distance concept and referring to the k-Nearest Neighbor concept [10], we can get the ratio of the nearest sections of the point in the projected dimensions.

**DEFINITION 4** (Nearest Sections in projected dimension)

Set point $p_{i,j}$, and project dimension i to k, where $p_j, p_f$ and $q \in section_i$. q is $p_j$'s any k-nearest neighbor points. $p_{k,f}$ is the point where $p_{i,f}$ and $p_{i,j}$ are in the same section. *s* is the number of points in the same section with $p_{i,j}$. Then *SecValp* of point $p_{i,j}$ is defined as follows:

$$SecValp(p_{i,j}, k) = \frac{dists(p_{k,j}, q))^2}{\frac{1}{s}\sum_{f=1}^{s} dists(p_{k,f}, q))^2} \quad (4)$$

We calculate the $p_j$'s *dists* with k-nearest neighbor points, and then get the ratio value with average value of points' *dists* in the section in the original dimension. While the dimension is projected to another dimension, it gets one *SecValp* value each time of projection. Totally, it gets $m\times(m-1)$ *SecValp* values from all the projected dimensions for each point.

3. Statistical Information Values for Each Point

Through the above two steps calculation, each point gets $m$ *SecVal* values in each dimension and gets $m\times(m-1)$ *SecValp* values in all the projected dimensions. We give the general statistical approach for each point $p_i$ in the section space, which is shown in Definition (5).

**DEFINITION 5** (Statistical Information of Point)

Set *SecVal*$(p_k)$ is the value of $p_k$ in each dimension, and *SecValp*$(p_k)$ is the value of $p_k$ in all projected dimensions. $\omega_1$ and $\omega_2$ are the weight for *SecVal* and *SecValp*. Then the statistical information value of $p_i$ is calculated as follows:

$$SI(p_k) = \omega_1 (\sum_{i=1}^{m} SecVal(p_k)) + \omega_2 (\sum_{i=1}^{m}\sum_{j=1, j\neq i}^{m} SecValp(p_k)) \quad (5)$$

SI (Statistic Information) is the point's value which is distinct among different points, and the outlier's SI value is obviously different from normal point's value. For the different dataset, changing the weight values could bring the better result. However, here, we use weight $\omega_1 = \frac{1}{m}$ and $\omega_2 = \frac{1}{m(m-1)}$ for clearly clarifying the SI.

The effective statistic method is considered in order to give the sharp boundary to compare with other points. By evaluating different methods and their performance, we choose a simple and clear calculation method. Here, we get the reciprocal value of average *SecVal* and *SecValp*. The outliers have obviously larger *SI* (Statistical Information) than that of the normal points'.

**DEFINITION 6** (Statistic Information of point)

$$SI(p_j) = \frac{2m}{\sum_{i=1}^{m}(SecVal(p_{i,j}))^2 + \frac{1}{m-1}\times\sum_{k=1}^{m-1} SecValp(p_{i,j},k)^2)} \quad (6)$$

This equation simply sums up *SecVal* and *SecValp* in all dimensions. Whether getting point's value in each dimension or in projected dimensions, all these point's values are the ratio with the average value in their respective dimensions. Most of the time normal points are around the average value in the dimension. Hence the *SI* value for normal points should be close to 1, and outlier's *SI* value should obviously larger than 1. However, it is not true in high dimensional space. Normal points' *SI* are getting smaller with the dimension increasing, and their values are always much lower than 1, as well as the outlier's values are also lower than 1. Nevertheless, the outlier's *SI* is still obviously higher than normal points'. Therefore, outliers can be easily found just by to sorting points by their *SI* values.

## 3.5 Algorithm

In this section, we focus on how to implement the k-NS method in R language. How to get *PointInfo* and *SectionInfo* effectively in different sections and dimensions is key issue that needs to be considered in detail. The proposed algorithm is shown in Figure 2 with pseudo-R code. Set the dataset containing $n$ points with $m$ dimension. The range of data is divided into *scn* sections in each dimension.

**Figure 2: k-NS Algorithm**

```
Algorithm: k-Nearest Section
Input: k, data[n,m], scn
Begin
Initialize(PointInfo[n, m], SectionInfo[scn, m])
For i=1 to m
  d_i=n/length(SectionInfo[ SectionInfo[i,]!=0,i])
  For j=1 to n
    Get PtVal_1[i,j] with Definition (3)
  End n
End m
For c=1 to 5
  Random sort dimension → i
  For i=1 to m
    For j=1 to scn
      PtNum <- SecInfo[j,i]
      If(PtNum ==0) next
      Ptid <- which(PtInfo[,i]==j)
      If(PtNum < 3/2 × k)  SecValp=1
      else  For each(p in Ptid])
      {
        if ( i<m) i=i+1
        else  i=1
        Get dists(P_{Ptid, i}) with Definition(1)
        Get SecValp with Definition(4)
      }
    End j
  End i
End c
Get SI value with Definition(6) for each point
Output: Outliers Points ID (SI(p) >> $\overline{SI}$ or top SI)
```

Three points need to be clarified in this algorithm.

The first point is how to decide the average section density $d_i$ in each dimension. $d_i$ value is easily obtained from calculation $\frac{n}{scn}$, which is the definition of the average section density. It means $d_i$ is same in each dimension. However, in the case that most points distribute in some small parts of sections and no point exists in other sections, $d_i$ becomes very low and even close to the outlier's section density. So, we only count sections with points, but don't count the sections without points for calculating $d_i$. Subsequently, $d_i$ reflects the average section density of points in $i^{th}$ dimension, but it also varies in different dimensions. Hence, the ratio of the section density to $d_i$ in Definition (3) can measure the sparsity of points in different sections in a dimension.

The second point is about the number of dimension loop in calculating all the point projections. For any point in a certain dimension, it can be projected to other *(m-1)* dimensions. Hence, the total dimension loops can be $m \times (m-1)$. However, the calculation cost of the all dimension loops is too expensive since the algorithm is applied to high dimensional space. Through the plenty of experiments, we have found that the number of projected dimension loop could be expressed by $\alpha \times m$ ($1 \leq \alpha \leq (m-1)$), and the proposed value for $\alpha$ is 5 where the projected dimension should be sorted by random order. Then the projected dimensions can be visited by the order from 1 to 2, 2 to 3, … , m to 1. The total dimension loop number is $5 \times m$ with a little accuracy loss, but it reduced the computing time greatly.

The third point is the number of points in one section. There are three different cases. Case 1: no point in the section. In this case, the algorithm just passes this section and goes to the next section. Case 2: many points in the section. In this case, the nearest section method is just used to detect points. Case 3: only a few points in the section. In this case, the point distribution is difficult to be judged just by these several points. And the section density ratio in the step 1 must be very low. Therefore, these points are to be already detected by the previous step. Here, we set the *SecValp*=1. The threshold value to separate the case 2 and case 3 is related to the *k*. *k* should not be large because *k* is less than $d_i$ in the projection dimension in the step 2. Through experiments with a serious of values from 4 to 20 to find the suitable value for *k* and the threshold of the number of points in one section, we have found that the threshold value can be defined as $\frac{3}{2} \times k$ as the best solution which could be adapted in most of the situations.

## 3.6 Distinction between Outliers and Noisy Data

The concept of outlier and noisy data has been proposed for more than ten years. According to that, outlier is regarded as abnormal data which is generated by a different mechanism and contains very important information, and noisy data are regarded as a side product of clustering points, which have no useful information but affect the correct result greatly.

In the data space, the outliers are points that are farther from others by some measures, while the noisy points always appear around the outliers. Since the noisy points are also far away from the normal points, in low dimensional space, it is difficult to make a distinct boundary between outliers and noisy points. Based on these frustrated observations, some researchers even consider that noisy data as a kind of outliers. There is no difference in detecting abnormal data by any methods. Hence, it is a meaningful issue to make a distinction between outliers and noisy points not only in concept but also in detection measures.

In this paper, we try to explain the distinction between outliers and noisy points in two aspects. The first point is that there are different data generation processes. Outliers are generated by a different distribution from normal points. Noisy points have the same distribution with normal points. The second point is that they appear abnormal in different dimensional space. Noisy points only appear abnormally in several dimensions and appear normal in other dimensions. From the whole dimensions' view, these noisy data also conform to the same distribution of normal data. The outlier may appear in the same way in low dimensional space, but they conform to different distribution mechanism from normal points'. Therefore, we can get a conclusion from these points that both outliers and noisy points can be detected in low dimensional space, but only outliers can be found in high dimensional space. An example of noisy data sample is shown in Figure 3. The data is retrieved from Dataset 8 as introduced in section 4, which contains 1000 points in 10000 dimensions. The outliers are placed in the middle region and can be found differently from normal points. The noisy points are labeled with cloud that is so different in this projected two dimensional space. Another example is shown in Figure 4. The outliers are not obvious in two-dimensional space, while noisy points that distribute on the marginal area of both dimensions are more likely abnormal points.

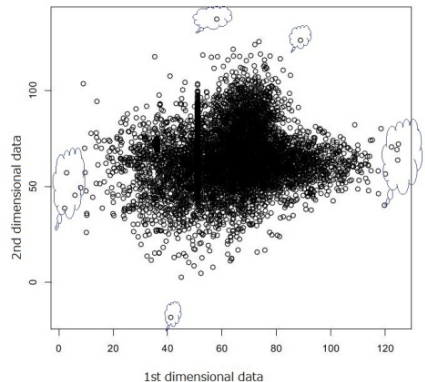

**Figure 3: Noisy Data of Dataset 3 Projected to Two-Dimensional Space**

## 4. EVALUATION

We have implemented our algorithm and applied it to several high dimensional datasets, and then have made the comparison among k-NS, LOF and our previous proposed methods. In order to compare these algorithms under fair conditions, we performed all algorithms with R language, on a Mac Book Pro with 2.53GHz Intel core 2 CPU and 4G memory.

On the issue of outlier detection in high dimensional space, we have proposed two algorithms, called RPGS (Rim Projected Grid Statistic) [6] and PSD (Projected Section Density) [under review]. RPGS is a statistic method that uses ratio of section density in each dimension and center section function in projected dimensions. The key definition about Center Section Function is shown in Definition 7. RPGS is only effective in one cluster data distribution in high dimensional space and cannot work well in more than two clustered data distributions.

**DEFINITION 7** (Center Section Function)

$$PtVal_2(x_i) = \frac{SecID(x_i) - Sec_{center}}{MaxSec - Sec_{center}} \quad (7)$$

$$PtVal_2(x_i) = \frac{Sec_{center} - SecID(x_i)}{Sec_{center} - MinSec} \quad (8)$$

*MinSec* is the minimum section ID; *MaxSec* is the maximum section ID. If $SecID > Sec_{center}$, Equation(7) is used, or else Equation (8) is used.

PSD is another proposed algorithm which detects outliers effectively in very high dimensional space. PSD is quite similar to k-NS algorithm, and it also includes the three-step detection method. PSD's peculiarity is on the second step for estimating outliers in projected dimension, which employs a new section-cluster concept, and calculate the ratio of section-cluster density to the mean value of points in the section. The calculation of section-cluster in PSD is shown in Definition (8).

**DEFINITION 8** (Section-cluster)

$$SecValp(p_{i,j},k) = \frac{Sec(p_{k,j}) \times CluLen_j}{\frac{1}{s}\sum_{f=1}^{s} Sec(p_{k,f}) \times CluLen_f} \quad (9)$$

Where the $CluLen_j$ is the cluster length according to the section of $p_j$ in $k$ dimension, $Sec(p_{k,j})$ is the number of points in the section where the point $p_j$ is projected in $k^{th}$ dimension from $i^{th}$ dimension by changing $k$, and $s$ is the number of points in the same section of $p_{i,j}$. The denominator is the average value of the points by the product of $Sec$ and the $CluLen$ in dimension $k$.

Compared with RPGS, PSD is an improved algorithm which can detect outlier in the case of different data distributions. In this paper, the normal points of test dataset conform to five clusters of Gaussian mixture model. RPGS performs poorly in the experiments. Therefore, we only show the experiment result of LOF, PSD and k-NS.

## 4.1 Synthetic Dataset

A critical issue of evaluating outlier detection algorithms is that no benchmark datasets have been found in real world to satisfy the explicit division between outliers and normal points. The points that are found as outliers in some real dataset are impossible to provide a reasonable explanation why these points are picked out as outliers. On the other hand, what we have learned from the statistical knowledge is helpful to generate the artificial dataset: if some points with some distributions are apparently different from those of normal points, these points can be regarded as outliers. Hence, we generate the synthetic data based on this assumption.

Since our algorithm aims to solve the problem in high dimensional space, we generate the eight synthetic datasets with points of 500-1000 and dimensions of 10-10000. The normal points conform to the normal distributions, while outliers conform to the random distributions in a fixed region. Normal points are equally distributed in five clusters and 10 outliers are distributed randomly in the middle of normal points' range. The more details about the parameters in each dataset are shown in Table 1.

**Table 1: Synthetic Dataset**

| Dataset No. | Dimension number | Points number | Normal points (Normal distribution) | Outliers Random |
|---|---|---|---|---|
| 1 | 10 | 500 | μ(20-80), σ(10-20) | (20-100) |
| 2 | 100 | 500 | μ(20-80), σ(10-20) | (20-100) |
| 3 | 100 | 1000 | μ(20-80), σ(10-20) | (20-100) |
| 4 | 500 | 500 | μ(20-80), σ(10-20) | (20-100) |
| 5 | 500 | 1000 | μ(20-80), σ(10-20) | (20-100) |
| 6 | 1000 | 500 | μ(20-80), σ(10-20) | (20-100) |
| 7 | 1000 | 1000 | μ(20-80), σ(10-20) | (20-100) |
| 8 | 10000 | 1000 | μ(20-80), σ(10-20) | (20-100) |

The synthetic datasets are generated by the special rules that outliers' range should be within the range of the normal points in any dimensions. Therefore, outliers cannot be found in low dimensional space. The data distribution example is shown in Figure 4 where the Dataset 3 is projected to two-dimensional space with outliers labeled with red color.

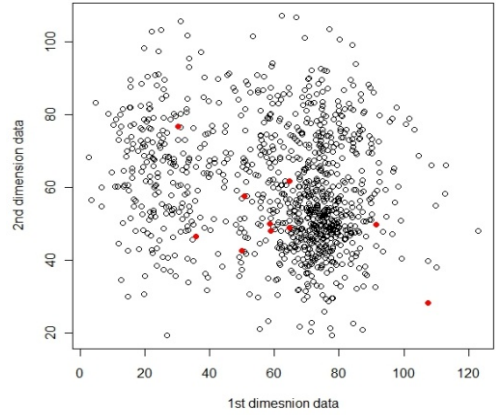

**Figure 4: Dataset 3 Projected to Two-Dimensional Space**

It is clearly shown in Figure 4 that the outliers are distributed within the range of normal points and show no difference to the normal points in this two-dimensional space. Noisy points that are placed on the margin of distributed area are more likely regarded as abnormal points. Hence, the different distributions for outliers and normal data cannot be found just by the straight observation. The outliers are designed to be loosely distributed, and they may be scattered further out of some fixed regions with small probability. Nevertheless these outliers are still within the range of normal points.

## 4.2 Effectiveness

In this subsection, our proposed algorithms is evaluated thoroughly by a series of experiments and compared with LOF and our previous proposed methods. In order to measure the performance of these algorithms to find most likely outliers at least false rate, the 10 outliers are repreived one by one. Therefore, it could provide the clear figures that describe the effectiveness of points are checked out on any extent of founded detected outliers. In the case of 10 dimensional dataset test, we use the precision and the recall for the better measurement of outlier detection efficiency among three algorithms. Since the recall is the percentage of all outliers in the dataset that have been found correctly, the 10 precision figures clearly show the trend on how effectively the outliers could be identified from all attained points. To contrast with the result affected by increasing the dimension from 10 to 100, we also use the precision and the recall measurement to evaluate the experiment results in the Dataset 2. In the processing of all eight dataset experiments, we evaluate the performance of the algorithms with F-measure by increasing dimension from 10 to 10000. In each dataset experiment, we obtained 10 precisions and the ten recall results respectively. Then the 10 F-measure values are obtained for each dataset test. We pick up the best F-measure from each dataset for better demonstrating the experiment performance by LOF, PSD and k-NS.

At the beginning, we need to set all the appropriate parameters for the three algorithms in eight experimental datasets, as shown in Table 2. The parameters shown in this table are the best ones for the prepared datasets, and they may changed according to the size of data, the number of dimensions, etc. The parameter Knn of

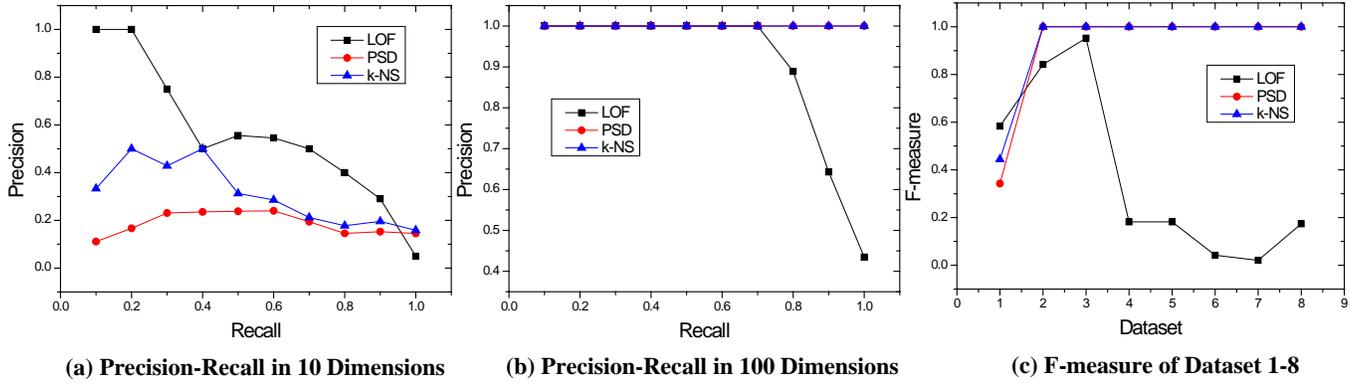

**(a) Precision-Recall in 10 Dimensions**    **(b) Precision-Recall in 100 Dimensions**    **(c) F-measure of Dataset 1-8**

**Figure 5: Effectiveness Comparison with LOF, PSD and k-NS in Eight Synthetic Datasets. Precision-Recall in 10 Dimensions and 100 Dimensions. F-measure Evaluation in Eight Datasets**

LOF is not set as a large value in all experiments because the dataset size is only around 500 or 1000 points. This is a reasonable ratio of neighbor points against the whole dataset size. For PSD algorithm, the parameters of *dst* and *scn* are inverse each other. The product of *dst* and *scn* is almost equal to the total points number. We set *scn* a little larger than *dst* because these combinations of parameters have shown the better experiment results. The parameters of k-NS are set similar to the previous two algorithms and are verified as applicable in all experiments.

**Table 2: Parameters of LOF, PSD and k-NS**

| Data ID | Points | Dimension | LOF | PSD | | k-NS | |
|---|---|---|---|---|---|---|---|
| | | | Knn | Dst | Scn | Knn | Scn |
| 1 | 500 | 10 | 8 | 20 | 25 | 5 | 25 |
| 2 | 500 | 100 | 10 | 20 | 25 | 6 | 25 |
| 3 | 1000 | 100 | 10 | 25 | 40 | 10 | 40 |
| 4 | 500 | 500 | 10 | 20 | 25 | 6 | 25 |
| 5 | 1000 | 500 | 10 | 30 | 34 | 10 | 34 |
| 6 | 500 | 1000 | 10 | 20 | 25 | 6 | 25 |
| 7 | 1000 | 1000 | 10 | 30 | 34 | 10 | 34 |
| 8 | 1000 | 10000 | 10 | 30 | 34 | 10 | 34 |

Since in our dataset design, the outliers are placed inside the range of normal data to prevent these points easily to be found in low dimensional space. For this reason, it is difficult to find exact outliers in 10 dimensional space. The 10 dimensional experiment result is shown in Figure 5(a). LOF performs best in this 10-dimensional test. Especially LOF can detect two outliers with very high precision. Nevertheless, the precision of LOF falls down sharply with the increasing recall from 20% to 40%. At last, the result of precision is worst among the three algorithms in detecting all outliers correctly. k-NS synthesize two algorithms' advantage, and the performance of k-NS is between LOF and PSD. To be noticed, the k-NS's precision is always higher than PSD at any recall rate.

We can get the conclusion from the experiment result as seen in Figure 5(a), the LOF gets high precision with low recall, gets low precision with high recall, and gets the worst precision with 100% recall. Both PSD and k-NS get low precision in low recall; get lower precision with the increased recall. The k-NS achieves double precision better than PSD's in low recall, and still better with increased recall.

When the number of dimension increases to 100, the precision and recall evaluation in 2$^{nd}$ dataset are clearly shown the effectiveness of three algorithms. Distinctly from the first dataset, both PSD and k-NS achieve the 100% precision with any recall all the time. LOF obviously reduces the precision from 100% to 43.48% with the increasing recall from 70% to 100%, as shown in Figure 5(b). In fact, PSD and k-NS keep the perfect result in 100 dimensions, while LOF performs much poorer with increasing the number of dimensions in terms of the precision and the recall.

To show the whole results of the effectiveness of three algorithms in all datasets with different number of dimensions, we also use the F-measure in each dataset. The F-measure can provide a more clear and simple evaluation index than the precision and the recall to contrast LOF, PSD and k-NS in different datasets.

In the experiment of datasets from 1 to 8, we evaluate the effectiveness for three algorithms as shown in Figure 5(c). LOF only achieves the best F-measure for each dataset, while PSD and k-NS only need to pick the largest F-measure in the first dataset. Since F-measures are always 1 from the Datasets 2 to 8.

It is clearly shown that PSD and k-NS perform perfectly in the number of dimensions from 100 to 10000 while LOF only works well in around 100 dimensional dataset. With the dimension increasing, LOF suffers the high dimensional curse greatly, and F-measure of LOF are less than 0.2 in the dataset 4-8. Learning from the dataset size test in Dataset 2 and 3, the three algorithms get better accuracy with the increasing dataset size.

### 4.3 Efficiencies

In this subsection, we compare three algorithms in running-time. In R language, the running time includes user time, system time and total time. So we only use the user time to compare them.

As shown in Figure 6, LOF is the fastest algorithm in all experiments. PSD runs a little faster than k-NS. The average running-time of K-NS is 1.51 times of PSD, and 6.65 times of LOF. The three algorithms take more time when the number of dimensions increases or the data size is enlarged. The reason is that there is no dimension-loop calculation for LOF because it only processes the distance between a point and its neighbors. However, our proposed algorithms PSD and k-NS both calculate values in all dimensions and in all different projected dimensions. Because k-NS gets calculation on neighbor points around a point, it takes more time to process than PSD.

### 4.4 Performance on Real World Data

In this subsection, we compare the three algorithms with a real-world dataset publicly available at the UCI machine learning repository. (http://archive.ics.uci.edu/ml/datasets/Arcene). We use Arcene dataset which is provided by ARCECE group.

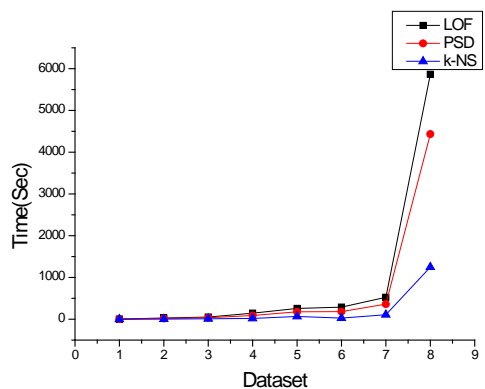

**Figure 6: Running Time**

The task of the group is to distinguish cancer versus normal patterns from mass-spectrometric data. This is a two-class classification problem with continuous input variables. This dataset is one of five datasets from the NIPS 2003 feature selection challenge.

The original dataset includes total 900 instances with 10000 attributes. The datasets have training dataset, validating dataset and test dataset. Each sub-dataset is labeled with positive and negative. In order to compare the algorithms clearly without distraction by unnecessary data, we only pick up positive test dataset (310 instances) as an evaluation target.

In the dataset, there are no true labeled outliers. Therefore, we evaluate the results with the top 20 value points, and try to find the number of the same points in these top 20 points.

The experimental result is shown in Table 3. The only one point (ID 310) is commonly detected by the three algorithms. The PSD and k-NS detect the common 16 points in their top 20 points, while LOF detects 19 different points. One possible reason is that PSD and k-NS use the same section-space framework while LOF uses the neighbor points' distance. Another possible reason is that LOF runs with poor performance in high dimensional space. The result also proves that LOF can find outliers, but in a limited way.

## 5. CONCLUSTIONS

In this paper, we introduce a new outlier detection method, called k-NS, being designed to efficiently detect outliers in a large and high dimensional dataset. The basic idea is the three-step statistical method, which (i) calculates the section density ratio in each dimension; (ii) computes the nearest sections ratio in all projected dimensions, and (iii) summarizes all values of each point for comparison with those of the other points. Our proposed k-NS algorithm has the following advantages:

- Immune to the high dimensional curse,
- Adapt to various outlier distributions,
- And outstanding performance in large scale dataset of high dimensional data space.

In the experimental evaluation we have demonstrated that k-NS performs significantly better than PSD in low dimensional space, achieves equally excellent results in high dimensional space. We also provide evidence of its effectiveness as compared with LOF in high dimensional space.

**Table 3: Top 20 points of arcane data**

| LOF SI | LOF ID | PSD SI | PSD ID | KNS SI | KNS ID |
|---|---|---|---|---|---|
| [1,] 2.431800 | 311 | [1,] 1.072082 | 312 | [1,] 0.06407881 | 189 |
| [2,] 1.702554 | 317 | [2,] 1.066246 | 237 | [2,] 0.06239231 | 221 |
| [3,] 1.479086 | 312 | [3,] 1.064365 | 310 | [3,] 0.06135894 | 310 |
| [4,] 1.404634 | 140 | [4,] 1.059466 | 132 | [4,] 0.06082506 | 132 |
| [5,] 1.341247 | 163 | [5,] 1.054983 | 221 | [5,] 0.05990445 | 237 |
| [6,] 1.338615 | 315 | [6,] 1.052735 | 190 | [6,] 0.05873731 | 190 |
| [7,] 1.317314 | 316 | [7,] 1.050393 | 34 | [7,] 0.05863396 | 312 |
| [8,] 1.236020 | 115 | [8,] 1.050011 | 50 | [8,] 0.05827137 | 34 |
| [9,] 1.211854 | 8 | [9,] 1.049966 | 189 | [9,] 0.05706454 | 50 |
| [10,] 1.203458 | 314 | [10,] 1.046804 | 192 | [10,] 0.05666884 | 242 |
| [11,] 1.198336 | 110 | [11,] 1.046405 | 180 | [11,] 0.05568051 | 78 |
| [12,] 1.170856 | 318 | [12,] 1.045346 | 78 | [12,] 0.05536066 | 180 |
| [13,] 1.139906 | 298 | [13,] 1.040544 | 126 | [13,] 0.05506420 | 111 |
| [14,] 1.130877 | 224 | [14,] 1.039460 | 152 | [14,] 0.05505178 | 48 |
| [15,] 1.127157 | 263 | [15,] 1.037209 | 108 | [15,] 0.05439610 | 131 |
| [16,] 1.121932 | 281 | [16,] 1.037118 | 92 | [16,] 0.05427036 | 29 |
| [17,] 1.121388 | 269 | [17,] 1.036678 | 159 | [17,] 0.05400753 | 192 |
| [18,] 1.119191 | 310 | [18,] 1.035495 | 29 | [18,] 0.05396448 | 307 |
| [19,] 1.114455 | 63 | [19,] 1.035128 | 111 | [19,] 0.05376915 | 126 |
| [20,] 1.114136 | 130 | [20,] 1.034683 | 242 | [20,] 0.05365066 | 92 |

The difference between outliers and noisy data is also discussed in this paper. This issue is difficult to be solved in low dimensional space since both data are always mixed together. In our experiments, the noisy data can be separated from outliers by projecting points from high dimensional space to two-dimensional space. The interesting point is that noisy data seem more abnormal than outliers in projected low dimensional space.

As the ongoing and future work, we continue to design and improve the algorithms under the section-space framework. More experiments need to be tested in order to seek for the perfect solution in outlier detection in high dimensional space. Another issue is the expensive cost of processing time in high dimensional data tests. Any solution to reduce the processing time needs to be investigated. One of the approaches may be the use of the parallel processing. This k-NS method can also be applied to other data mining technology such as clustering in high dimensions, classifications, etc.